# Maximized Posteriori Attributes Selection from Facial Salient Landmarks for Face Recognition

Phalguni Gupta[1], Dakshina Ranjan Kisku[2], Jamuna Kanta Sing[3], Massimo Tistarelli[4]

[1] Department of Computer Science and Engineering,
Indian Institute of Technology Kanpur,
Kanpur - 208016, India
[2] Department of Computer Science and Engineering,
Dr. B. C. Roy Engineering College / Jadavpur University,
Durgapur – 713206, India
[3] Department of Computer Science and Engineering,
Jadavpur University, Kolkata – 700032, India
[4] Computer Vision Laboratory, DAP
University of Sassari, Alghero (SS), 07041, Italy
{drkisku, jksing}@ieee.org; pg@cse.iitk.ac.in; tista@uniss.it

**Abstract.** This paper presents a robust and dynamic face recognition technique based on the extraction and matching of devised probabilistic graphs drawn on SIFT features related to independent face areas. The face matching strategy is based on matching individual salient facial graph characterized by SIFT features as connected to facial landmarks such as the eyes and the mouth. In order to reduce the face matching errors, the Dempster-Shafer decision theory is applied to fuse the individual matching scores obtained from each pair of salient facial features. The proposed algorithm is evaluated with the ORL and the IITK face databases. The experimental results demonstrate the effectiveness and potential of the proposed face recognition technique also in case of partially occluded faces.

## 1 Introduction

Face recognition can be considered as one of most dynamic and complex research areas in machine vision and pattern recognition [1,2] because of the variable appearance of face images. The appearance changes in face occur either due to intrinsic and extrinsic factors and due to these changes, face recognition problems become ill posed and difficult to authenticate faces with outmost ease. Auxiliary complexities like the facial attributes compatibility complexity, data dimensionality problem, the motion of face parts, facial expression changes, pose changes, partly

2      **Phalguni Gupta1, Dakshina Ranjan Kisku2, Jamuna Kanta Sing3,** Massimo Tistarelli4

occlusion and illumination changes cause major changes in appearance. In order to make the problem well-posed, vision researchers have adapted and applied an abundance of algorithms for pattern classification, recognition and learning.

There exist the appearance-based techniques which include Principal Component Analysis (PCA) [1], Linear Discriminant Analysis (LDA) [1], Fisher Discriminant Analysis (FDA) [1] and Independent Component Analysis (ICA) [1]. Some local feature based methods are also investigated [4-5]. A local feature-based technique for face recognition, called Elastic Bunch Graph Matching (EBGM) has been proposed in [3]. EBGM is used to represent faces as graphs and the vertices localized at fiducial points (e.g., eyes, nose) and the geometric distances or edges labeled with the distances between the vertices. Each vertex contains a set known as Gabor Jet, of 40 complex Gabor wavelet coefficients at different scales and orientations. In case of identification, these constructed graphs are searched and get one face that maximizes the graph similarity function. There exists another graph-based technique in [6] which performs face recognition and identification by graph matching topology drawn on SIFT features [7-8]. Since the SIFT features are invariant to rotation, scaling and translation, the face projections are represented by graphs and faces can be matched onto new face by maximizing a similarity function taking into account spatial distortions and the similarities of the local features.

This paper addresses the problem of capturing the face variations in terms of face characteristics by incorporating probabilistic graphs drawn on SIFT features extracted from dynamic (mouth) and static (eyes, nose) salient facial parts. Differences in facial expression, head pose changes, illumination changes, and partly occlusion, result variations in facial characteristics and attributes. Therefore, to combat with these problems, invariant feature descriptor SIFT is used for the proposed graph matching algorithm for face recognition which is devised pair-wise manner to salient facial parts (e.g., eyes, mouth, nose).

The goal of the proposed algorithm is to perform an efficient and cost effective face recognition by matching probabilistic graph drawn on SIFT features whereas the SIFT features [7] are extracted from local salient parts of face images and directly related to the face geometry. In this regard, a face-matching technique, based on locally derived graph on facial landmarks (e.g., eye, nose, mouth) is presented with the fusion of graphs in terms of the fusion of salient features. In the local matching strategy, SIFT keypoint features are extracted from face images in the areas corresponding to facial landmarks such as eyes, nose and mouth. Facial landmarks are automatically located by means of a standard facial landmark detection algorithm [8-9]. Then matching a pair of graphs drawn on SIFT features is performed by searching a most probable pair of probabilistic graphs from a pair of salient landmarks. This paper also proposes a local fusion approach where the matching scores obtained from each pair of salient features are fused together using the Dempster-Shafer decision theory. The proposed technique is evaluated with two face databases, viz. the IIT Kanpur and the ORL (formerly known as AT&T) databases [11] and the results demonstrate the effectiveness of the proposed system.

The paper is organized as follows. The next section discusses SIFT features extraction and probabilistic graph matching for face recognition. Experimental results are presented in Section 3 and conclusion is given in the last section.

## 2   SIFT Feature Extraction and Probabilistic Matching

### 2.1   SIFT Keypoint Descriptor for Representation

The basic idea of the SIFT descriptor [6-7] is detecting feature points efficiently through a staged filtering approach that identifies stable points in the scale-space. Local feature points are extracted by searching peaks in the scale-space from a difference of Gaussian (DoG) function. The feature points are localized using the measurement of their stability and orientations are assigned based on local image properties. Finally, the feature descriptors which represent local shape distortions and illumination changes, are determined.

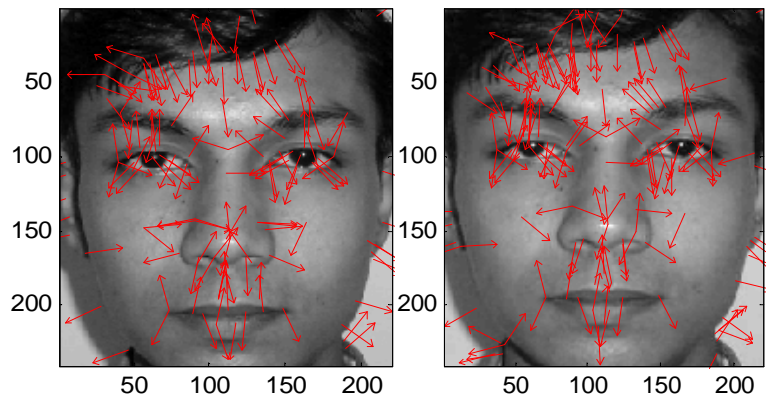

**Fig. 1.** Invariant SIFT Feature Extraction on a pair of Face Images

Each feature point is composed of four types of information – spatial location ($x$, $y$), scale ($S$), orientation ($\theta$) and Keypoint descriptor ($K$). For the sake of the experimental evaluation, only the keypoint descriptor [6-7] has been taken into account. This descriptor consists of a vector of 128 elements representing the orientations within a local neighborhood. In Figure 1, the SIFT features extracted from a pair of face images are shown.

### 2.2   Local Salient Landmarks Representation with Keypoint Features

Deformable objects are generally difficult to characterize with a rigid representation in feature spaces for recognition. With a large view of physiological characteristics in biometrics including iris, fingerprint, hand geometry, etc, faces are considered as highly deformable objects. Different facial regions, not only convey different relevant and redundant information on the subject's identity, but also suffer from different time variability either due to motion or illumination changes. A typical example is the case of a talking face where the mouth part can be considered as dynamic facial landmark part. Again the eyes and nose can be considered as the static facial landmark parts which are almost still and invariant over time. Moreover, the mouth moves changing its appearance over time. As a consequence, the features extracted from the mouth area cannot be directly matched with the corresponding features from a static template. Moreover, single facial features may be occluded making the corresponding image area not usable for identification. For these reasons to improve the identification and recognition process, a method is performed which searches the matching features from a pair of facial landmarks correspond to a pair of faces by maximizing the posteriori probability among the keypoints features. The aim of the proposed matching technique is to correlate the extracted SIFT features with independent facial landmarks. The SIFT descriptors are extracted and grouped together by searching the sub-graph attributes and drawing the graphs at locations corresponding to static (eyes, nose) and dynamic (mouth) facial positions.

The eyes and mouth positions are automatically located by applying the technique proposed in [8]. The position of nostrils is automatically located by applying the technique proposed in [9]. A circular region of interest (ROI), centered at each extracted facial landmark location, is defined to determine the SIFT features to be considered as belonging to each face area.

SIFT feature points are then extracted from these four regions and gathered together into four groups. Then another four groups are formed by searching the corresponding keypoints using iterative relaxation algorithm by establishing relational probabilistic graphs [12] on the four salient landmarks of probe face.

### 2.3   Probabilistic Interpretation of Facial Landmarks

In order to interpret the facial landmarks with invariant SIFT points and probabilistic graphs, each extracted feature can be thought as a node and the relationship between invariant points can be considered as geometric distance between the nodes. At the level of feature extraction, invariant SIFT feature points are extracted from the face images and the facial landmarks are localized using the landmark detection algorithms discussed in [8], [9]. These facial landmarks are used to define probabilistic graph which is further used to make correspondence and matching between two faces.

To measure the similarity of vertices and edges (geometric distances) for a pair of graphs [12] drawn on two different facial landmarks of a pair of faces, we need to measure the similarity for node and edge attributes correspond to keypoint descriptors and geometric relationship attributes among the keypoints features. Let, two graphs be $G' = \{N', E', K', \varsigma'\}$ and $G'' = \{N'', E'', K'', \varsigma''\}$ where $N'$, $E'$, $K'$, $\varsigma'$ denote nodes, edges, association between nodes and association between edges respectively. Therefore, we can denote the similarity measure for nodes $n'_i \in N'$ and $n''_j \in N''$ by $s^n_{ij} = s(k'_i, k''_j)$ and the similarity between edges $e'_{ip} \in E'$ and $e''_{jq} \in E''$ can be denoted by $s^e_{ipjq} = s(e'_{ip}, e''_{jq})$.

Further, suppose, $n'_i$ and $n''_j$ are vertices in gallery graph and probe graph, respectively. Now, $n''_j$ would be best probable match for $n'_i$ when $n''_j$ maximizes the posteriori probability [12] of labeling. Thus for the vertex $n'_i \in N'$, we are searching the most probable label or vertex $\overline{n}'_i = n''_j \in N''$ in the probe graph. Hence, it can be stated as

$$\overline{n}'_i = \arg \max_{j, n''_j \in N''} P(\psi_i^{n''_j} \mid K', \varsigma', K'', \varsigma'') \tag{1}$$

To simplify the solution of matching problem, we adopt a relaxation technique that efficiently searching the matching probabilities $\overline{P}^n_{ij}$ for vertices $n'_i \in N'$ and $n''_j \in N''$. By reformulating Equation (1) can be written as

$$\overline{n}'_i = \arg \max_{j, n''_j \in N''} \overline{P}^n_{ij} \tag{2}$$

This relaxation procedure considers as an iterative algorithm for searching the best labels for $\overline{n}'_i$. This can be achieved by assigning prior probabilities $P^n_{ij}$ proportional to $s^n_{ij} = s^n(k'_i, k''_j)$. Then the iterative relaxation rule would be

$$\hat{P}^n_{ij} = \frac{P^n_{ij} . Q_{ij}}{\sum_{q, n''_q \in N''} P^n_{iq} . Q_{iq}} \tag{3}$$

$$Q_{ij} = p^n_{ij} \prod_{p, n'_p \in N'_i} \sum_{q, n''_q \in N''_j} s^e_{ipjq} . P^n_{pq} \tag{4}$$

Relaxation cycles are repeated until the difference between prior probabilities $P^n_{ij}$ and posteriori probabilities $\hat{P}^n_{ij}$ becomes smaller than certain threshold $\Phi$ and when this is reached then it is assumed that the relaxation process is stable. Therefore,

$$\max_{i, n'_i \in N', j, n''_j \in N''} \left| P^n_{ij} - \hat{P}^n_{ij} \right| < \Phi \tag{5}$$

Hence, the matching between a pair of graphs is established by using the posteriori probabilities in Equation (2) about assigning the labels from the gallery graph $G'$ to the points on the probe graph $G''$.

From these groups pair-wise salient feature matching is performed in terms of graph matching. Finally, the matching scores obtained from these group pairs are

6      Phalguni Gupta1, Dakshina Ranjan Kisku2, Jamuna Kanta Sing3, Massimo Tistarelli4

fused together by the Dempster-Shafer fusion rule [10] and the fused score is compared against a threshold for final decision.

## 3   Experimental Evaluation

To investigate the effectiveness and robustness of the proposed graph-based face matching strategy, experiments are carried out on the IITK face database and the ORL face database [11]. The IITK face database consists of 1200 face images with four images per person (300X4), which have captured in control environment with ±20 degree changes of head pose and with almost uniform lighting and illumination conditions, and the facial expressions keeping consistent with some ignorable changes. For the face matching, all probe images are matched against all target images. On the other hand, the ORL face database consists of 400 images taken from 40 subjects. Out of these 400 images, we use 200 face images for experiment, in which ±20 to ±30 degrees orientation changes have been considered. The face images show variations of pose and facial expression (smile/not smile, open/closed eyes). When the faces have been taken, the original resolution is 92 x 112 pixels for each one. However, for our experiment we set the resolution as 140×100 pixels in line with IITK database.

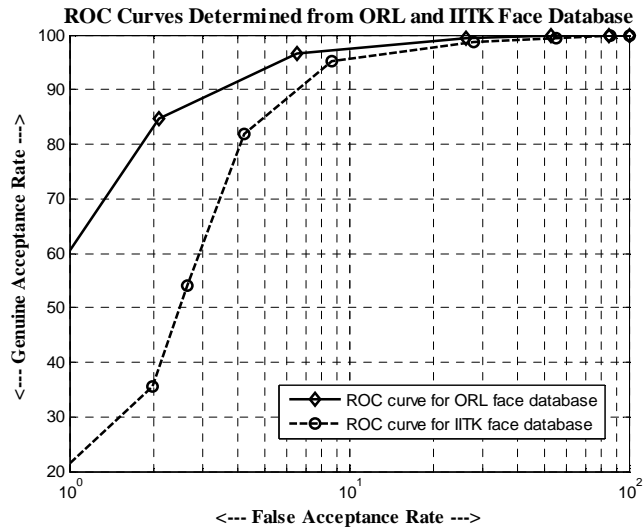

**Fig. 2.** ROC curves for the proposed matching algorithm for ORL and IITK databases.

   The ROC curves of the error rates obtained from the face matching applied to the face databases are shown in Figure 2. The computed recognition accuracy for the IITK database is 93.63% and for the ORL database is 97.33%. The relative accuracy of the proposed matching strategy for ORL database increases of about 3% over the IITK database. In order to verify the effectiveness of the proposed face matching algorithm for recognition and identification, we compare our algorithm with the algorithms that are discussed in [6], [13], [14], and [15]. There are several face matching algorithms discussed in the literatures which tested on different face databases or with different processes. It is duly unavailable of such uniform experimental environment, where the experiments can be performed with multiple attributes and characteristics. By extensive comparison, we have found that, the proposed algorithm is solely different from the algorithms in [6], [13], [14], [15] in terms of performance and design issues. In [13], the PCA approach discussed for different view of face images without transformation and the algorithm achieved 90% recognition accuracy for some specific views of faces. On the other hand, [14] and [15] use Gabor jets for face processing and recognition where the first one has used the Gabor jets without transformation and later one has used the Gabor jets with geometrical transformation. Both the techniques are tested on Bochum and FERET databases which are characteristically different from the IITK and the ORL face databases and the recognition rates are 94% and 96%, respectively at maximum, while all the possible testing are done with different recognition rates. Also, another two graph based face recognition techniques drawn on SIFT features have been discussed in [6] where the graph matching algorithms are developed by considering the whole face instead of the local landmark areas. The proposed face recognition algorithm not only devised keypoints from the local landmarks, but it also combines the local features for robust performance.

## 4   Conclusion

This paper has proposed an efficient and robust face recognition techniques by considering facial landmarks and using the probabilistic graphs drawn on SIFT feature points. During the face recognition process, the human faces are characterized on the basis of local salient landmark features (e.g., eyes, mouth, nose). It has been determined that when the face matching accomplishes with the whole face region, the global features (whole face) are easy to capture and they are generally less discriminative than localized features. On contrary, local features on the face can be highly discriminative, but may suffer for local changes in the facial appearance or partial face occlusion. In the proposed face recognition method, local facial landmarks are considered for further processing rather than global features. The optimal face representation using probabilistic graphs drawn on local landmarks allow matching the localized facial features efficiently by searching and making correspondence of keypoints using iterative relaxation by keeping similarity measurement intact for face recognition.

8     **Phalguni Gupta1, Dakshina Ranjan Kisku2, Jamuna Kanta Sing3,** Massimo Tistarelli4